\documentclass[10pt,twocolumn,letterpaper]{article}

\usepackage{cvpr}
\usepackage{times}
\usepackage{epsfig}
\usepackage{graphicx}
\usepackage{amsmath}
\usepackage{amssymb}

\usepackage{multirow}
\usepackage{array}
\usepackage{color}
\usepackage{subfigure}
\usepackage{colortbl}
\usepackage{arydshln}
\graphicspath{{images/}} 

\usepackage[pagebackref=true,breaklinks=true,letterpaper=true,colorlinks,bookmarks=false]{hyperref}

\cvprfinalcopy 

\def\cvprPaperID{7934} 

\ifcvprfinal\fi
\begin{document}

\title{Macroscopic Control of Text Generation for Image Captioning }

\author{Zhangzi Zhu, Tianlei Wang, and Hong Qu\\
University of Electronic Science and Technology of China\\
{\tt\small 202021080414@std.uestc.edu.cn}
}

\maketitle

\begin{abstract}
   Despite the fact that image captioning models have been able to generate impressive descriptions for a given image, challenges remain: (1) the controllability and diversity of existing models are still far from satisfactory; (2) models sometimes may produce extremely poor-quality captions. In this paper, two novel methods are introduced to solve the problems respectively. Specifically, for the former problem, we introduce a control signal which can control the macroscopic sentence attributes, such as sentence quality, sentence length, sentence tense and number of nouns etc. With such a control signal, the controllability and diversity of existing captioning models are enhanced. For the latter problem, we innovatively propose a strategy that an image-text matching model is trained to measure the quality of sentences generated in both forward and backward directions and finally choose the better one. As a result, this strategy can effectively reduce the proportion of poor-quality sentences. Our proposed methods can be easily applied on most image captioning models to improve their overall performance. Based on the Up-Down model, the experimental results show that our methods achieve BLEU-4/CIDEr/SPICE scores of 37.5/120.3/21.5 on MSCOCO Karpathy test split with cross-entropy training, which surpass the results of other state-of-the-art methods trained by cross-entropy loss.
\end{abstract}

\begin{table}[ht]
	\begin{center}
		\begin{tabular}{|c|p{80 pt}|m{10 pt}<{\centering}|m{19 pt}<{\centering}|m{8 pt}<{\centering}|}
			\hline
			image & captions & len & tense & n \\
			\hline

			\multirow{2}{*}[-15 pt]{\includegraphics[width=60pt]{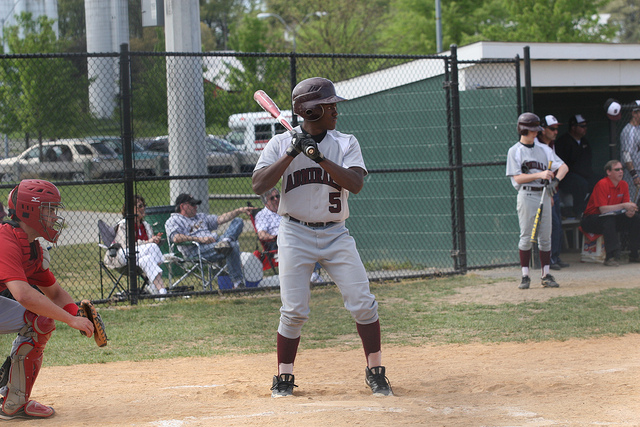}} & Base: a \textcolor{blue}{baseball} \textcolor{cyan}{player} \textit{holding} a \textcolor{green}{bat} on a \textcolor{magenta}{field}. & \multirow{1}{*}[-15 pt]{9} & \multirow{1}{*}[-15 pt]{v-ing} & \multirow{1}{*}[-15 pt]{4} \\
			\cline{2-5}
			& \textbf{Ours:} a \textcolor{blue}{baseball} \textcolor{cyan}{player} \underline{\textit{holds}} a red \textcolor{green}{bat} on a \textcolor{magenta}{field}. & \multirow{1}{*}[-15 pt]{10} & \multirow{1}{*}[-15 pt]{v} & \multirow{1}{*}[-15 pt]{4} \\
			\hline
			
			\multirow{3}{*}[-20 pt]{\includegraphics[width=60pt]{COCO_val2014_000000216228.jpg}} & Base: a \textcolor{cyan}{women} \textit{riding} a \textcolor{magenta}{skateboard} down a \textcolor{green}{street}. & \multirow{1}{*}[-15 pt]{8} & \multirow{1}{*}[-15 pt]{v-ing} & \multirow{1}{*}[-15 pt]{3} \\
			\cline{2-5}
			& \textbf{Ours:} a \textcolor{cyan}{women} \underline{\textit{rides}} a \textcolor{magenta}{skateboard} down a \textcolor{green}{street} near \textcolor{blue}{shops}. & \multirow{1}{*}[-15 pt]{10} & \multirow{1}{*}[-15 pt]{v} & \multirow{1}{*}[-15 pt]{4} \\
			\hline

			\multirow{3}{*}[-15 pt]{\includegraphics[width=60pt]{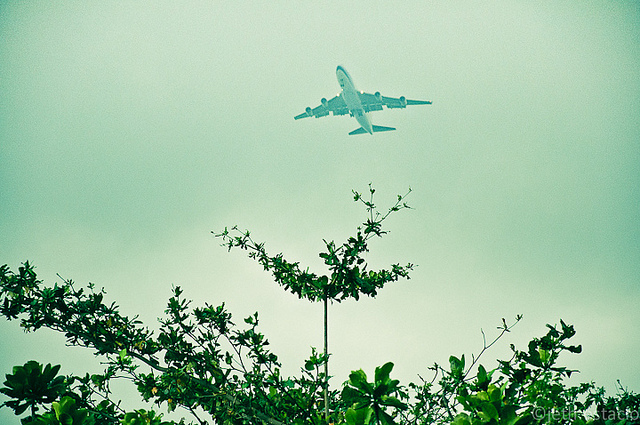}}& Base: a \textcolor{cyan}{plane} \textit{flies} in the \textcolor{magenta}{sky} above a \textcolor{green}{tree} & \multirow{1}{*}[-15 pt]{9} & \multirow{1}{*}[-15 pt]{v} & \multirow{1}{*}[-15 pt]{3} \\
			\cline{2-5}
			& \textbf{Ours:} an \textcolor{cyan}{airplane} \underline{\textit{flies}} in the \textcolor{magenta}{sky} with \textcolor{green}{trees} in \textcolor{blue}{background} & \multirow{1}{*}[-15 pt]{10} & \multirow{1}{*}[-15 pt]{v} & \multirow{1}{*}[-15 pt]{4} \\
			\hline
		\end{tabular}
	\end{center}
	\caption{Example images with captions generated by base\cite{Anderson2018Bottom} and our method. As shown, a description of specific requirements such as sentence length of 10, present tense, and 4 nouns contained is needed. The base model fails to be controlled, but our control method can fully meet the needs.}
\label{Table1}\end{table}

\section{Introduction}

Image caption, which belongs to the intersection of computer vision and natural language processing, is an important part of applying artificial intelligence to many life scenes. The generated texts can be used for image search task and helping the visually impaired. Caption generation is a particularly challenging task which requires models not only to recognize salient objects, attributes and relationships in an image, but also to describe various information through fluent natural language.

In the past few years, most captioning models \cite{Dai2018Rethinking, Qin2020Look, ChenRegularizing, Zhou2020MoreGI} followed encoder-decoder framework which is proposed in \cite{Vinyals2015Show}. In such a framework for image captioning, encoder and decoder are composed of Convolution Neural Network(CNNs) and Recurrent Neural Network(RNNs) respectively. The encoder is used to extract the features of the image, and then the decoder generates corresponding sequence sentences based on the image feature information. Attention mechanism is first introduced into this task in \cite{2015Show}, which enables models to quickly scan the global image at each time step, selectively get the target area of interest that needs special attention. Visual attention mechanisms have brought significant improvement in captioning and been widely adopted by recent models.

Despite the efforts made for captioning task, there still exist two problems. One problem is that existing captioning models lack controllability and diversity. The existing model acts like a black box. It takes a given picture as input and outputs a natural language sentence, which cannot be influenced by the exterior. By contrast, humans can generate personalized descriptions of a picture according to the needs of the task. When faced with some external constraints, they are always able to intelligently generate qualified sentences for specific needs. The lack of controllability creates a huge gap between human and machine intelligence.

The other problem is that no matter how good a model is, it will generate particularly poor-quality sentences when confronted with some images in the dataset. These poor-quality sentences will have a negative effect in real life. Therefore, it is crucial to find a way to reduce the proportion of bad sentence generation.

In this paper, we introduce two effective methods to remedy the above problem. Specifically, we propose a control signal whose form is a continuous value or a discrete value. It provides a way for caption generation to be externally controlled. As such, our method is capable of describing the same image with various sentence attributes, following the given conditioning. Table~\ref{Table1} shows a comparison between models with or without the control signal. It can be seen that model with the control signal is able to generate more controllable sentences which can fully meet the needs from the exterior. To the best of our knowledge, this is the first image captioning method through which people can accurately control the macroscopic sentence attributes like sentence quality, sentence length, sentence tense, or number of nouns in the sentence. To handle the second problem, we provide an extra option of caption for each image. For a given image, both two different captioning models are much less likely to generate extremely poor-quality sentences than a single model. Following this idea, we train two captioning models, one to generate sentences from front to back and the other to reverse. Then we additionally train an image-text matching model serving as an evaluator to choose the better one between two sentences. Experiments conducted on MSCOCO dataset show that this method can effectively reduce the proportion of poor-quality sentences and improve overall captioning performance.

As above, we mainly have the following contributions in this paper:

(1).	We introduce the control signal. Through it, macroscopic sentence attributes like sentence quality, sentence length, sentence tense, or number of nouns in the sentence can be controlled from the exterior, which enhances the controllability and diversity of existing captioning models. 

(2).	We innovatively train an image-text matching model to judge the quality of sentences generated in forward and backward directions and choose the better one, which can effectively reduce the proportion of poor-quality sentences and improve overall captioning performance.

(3).	By combining two methods, we achieve a new state-of-the-art performance on MSCOCO dataset in terms of cross-entropy training. With CIDEr optimization, our methods also make huge improvement compared with the baseline. Besides, our methods are generic, which can be applied to most captioning models.

\section{Related Work}

\begin{figure*}[htbp]
	\begin{center}
		\includegraphics[scale=0.3]{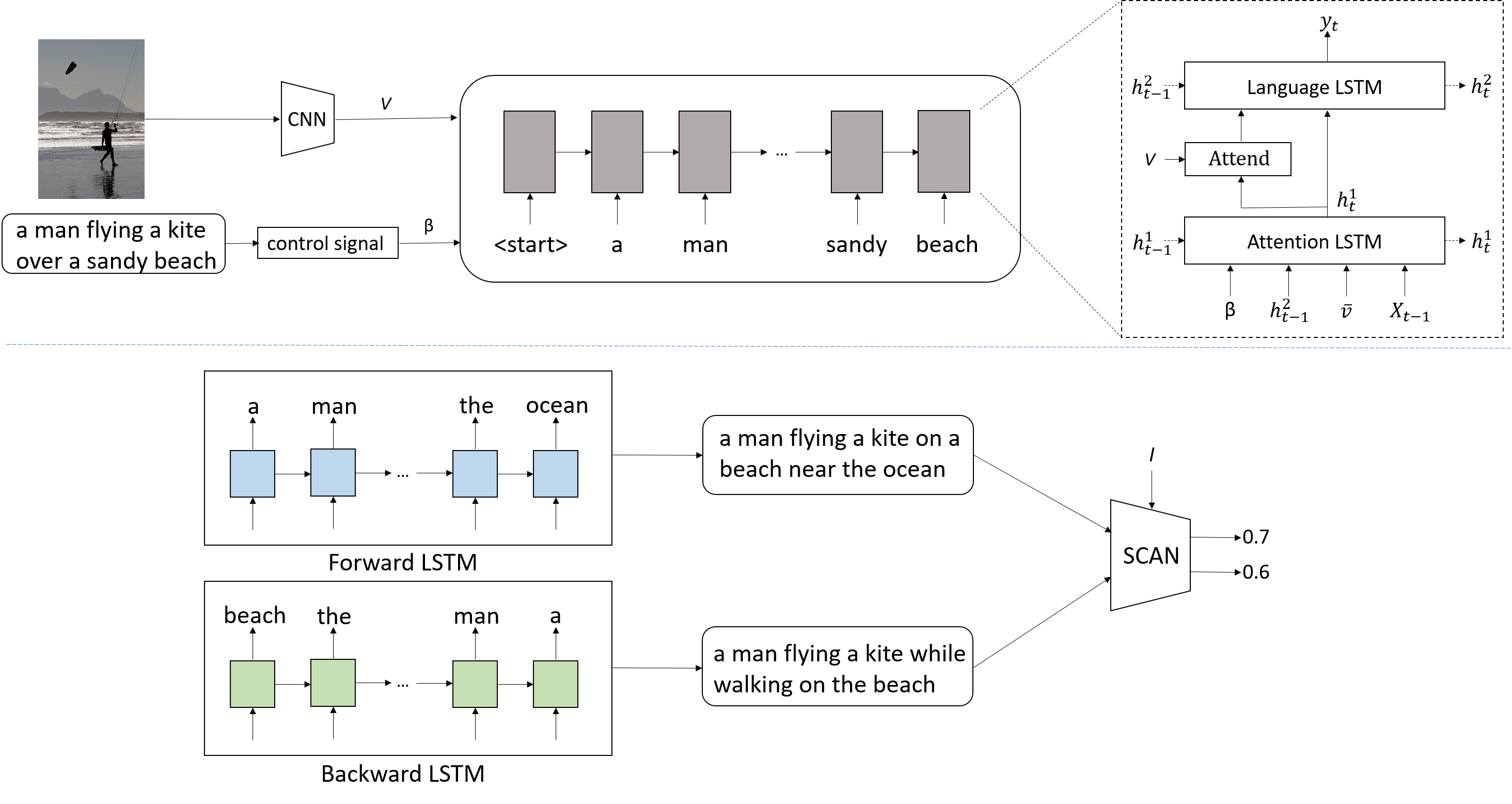}
	\end{center}
	\caption{Overview of our methods. Top: Given pairs of images and control signals, we first train two caption models (Forward LSTM and Backward LSTM) with cross-entropy loss and CIDEr optimization. Bottom: We then train an image-text matching model to choose the better one which is more alignment to the image between sentences generated by two caption models above. }
	\label{fig:overview}
\label{Figure1}\end{figure*}

\subsection{Image Caption}

In the early period, captioning models first generate slotted caption templates and then fill in the slots with specific words, which are named as rule/template-based models. Recently, most captioning models follow an encoder-decoder framework which is first introduced to image captioning task by Show-Tell model \cite{Vinyals2015Show}. In \cite{Huang2019Attention, 2016Image, 2017Knowing, Yang2019Learning, Long2017SCA, Wang2020Show}, attention mechanism is used to enable the model to selectively get the target area of interest that needs special attention at each time step. In \cite{Rennie2016Self, Dognin2019Adversarial}, reinforce learning method is applied to optimize non-differentiable metrics, which solves the problem of exposure bias. In \cite{Dognin2019Adversarial, Dai2017Towards, Chen2018Improving}, generative adversarial network (GAN) is introduced in order to enhance the diversity of sentences. Models proposed in \cite{Zheng2019Intention, Ge2019Exploring} change the order of the sentence generation, starting from the middle or the end of the sentence. In \cite{Yang2018Auto, Chen2020Say, shi-etal-2020-improving}, scene graphs are employed to futher explore the objects, attributes and relationships in the image, which improve the overall performance of captioning models.

In addition to the accuracy and diversity, another related route is to generate controllable sentences. Models in \cite{Zheng2019Intention} generate sentences starting from a guiding object in order to contain the given word. Methods in \cite{Cornia2020Show} describe images conditioning on a given sequence or set of image regions, which can control which objects are described and their orders. Models in \cite{Deshpande2019Fast} condition sentence probabilities on a Part-of-Speech (POS) tag sequence to generate the corresponding sentence. Models in \cite{Chen2020Say} control sentences at a more fine-grained level, which take Abstract Scene Graphs (ASG) as the control signal.

All above works concentrate on the fine-grained level such as specific objects, orders of objects or part-of-speech at each time step. However, they ignore the macroscopic sentence attributes. In this paper, we propose a control signal in the form of a continuous value or a discrete value to control the macroscopic sentence attributes such as sentence quality, sentence length, sentence tense, or number of nouns in the sentence, which enhances the controllability at a macro level.

\subsection{Image-Text Matching}

Existing image-text matching models can be divided into two categories: global matching and local matching. Global matching based methods \cite{2017Learning, Gu2018Look, Wang2017Learning, Wu2018Learning} embed the whole image and sentence to a joint space and calculate the similarity between them. Local matching based methods \cite{Lee2018Stacked,Huang2018Bi, Wang2019Position} infer the similarity between the local region and word. In this paper, we employ the classic local matching model SCAN \cite{Lee2018Stacked} to act as an evaluator to choose the one which is more alignment to the image between two different sentences.

\section{Approch}

As illustrated in Figure~\ref{Figure1}, our model consists of two image caption models and an image-text matching model. We will first introduce the specific structure of the two components respectively, and then elaborate on how to combine the two components to generate more controllable and high-quality sentences.

\subsection{Image Caption Model}

In this paper, we take the classic Up-Down model proposed by \cite{Anderson2018Bottom} as our base model since its remarkable performance. It consists of a CNN-based encoder and a LSTM-based decoder with the attention module. Given an image I, we first generate a set of regions feature $V=\{v_1,v_2,…,v_k\}$, $v_i\in\mathbb{R}^D $extracted by a detector\cite{Ren2015Faster}, which is then transformed as:

\begin{equation}
	v_i = w_v f_i + b_v.
\end{equation}

Then at each step, the input of the attention LSTM is a concatenated vector of the control signal $ \beta $, the previous output of the language LSTM $ h_{t-1}^2 $, the global pooling vector $ \overline{v} = \begin{matrix} \tfrac{1}{k} \sum_{k=1} v_i \end{matrix} $ and previous word embedding $ X_{t-1} $:

\begin{equation}
	h_t^1 = LSTM_1([\beta;h_{t-1}^1;\overline{v};X_{t-1}],h_{t-1}^1),
\end{equation}

\noindent where $ \beta \in \mathbb{R} $ is a signal which can make the sentence more controllable. To generate the high-quality sentences, you can set the value of $ \beta $ to the cider score\cite{Vedantam2015CIDEr} of the label sentence, which is calculated between the label sentence and five ground truth. Also, you can set it to a discrete value for other demand, depending on the category index of sentence length, sentence tense or number of nouns in the sentence. If you want to control four sentence attributes simultaneously in one model, the dimension of $ \beta $ can be expanded to 4 and each dimension is able to control one of the sentence attributes. Given the output $ h_t^1 $ of the attention LSTM, The attended image feature used as input to the language LSTM is calculated as follows:

\begin{equation}
	z_{i,t} = W_\alpha^T tanh(W_{v\alpha}v_i + W_{h\alpha}h_t^\prime),
\end{equation}

\begin{equation}
	\begin{aligned}
		\hat{v}_t = \sum_{i=1}^K \alpha_{i,t}v_i, \quad \alpha_i = softmax(z_t).
	\end{aligned}
\end{equation}

The concatenated vector of the attended image feature $ v_t $ and $ h_t^1 $ are then fed into the language LSTM, which finally outputs the conditional distribution over possible output word as:

\begin{equation}
	h_t^2 = LSTM_2([\hat{v}_t;h_t^1],h_{t-1}^2),
\end{equation}

\begin{equation}
	p(y_t|y_{1:t-1}) = softmax(W_ph_t^2 + b_p).
\end{equation}

\subsection{Image-Text Matching Model}
In this paper, we adopt the classic image-text matching model SCAN \cite{Lee2018Stacked} to act as an evaluator. It can infer the image-sentence similarity by aligning image region and word features. By using this, we can choose the one which is more alignment to the image between two different sentences generated in forward and backward directions. Here, we will briefly introduce SCAN model. Specifically, given an image I and a sentence T, we first use Faster-RCNN \cite{Ren2015Faster} and a bi-directional GRU \cite{Schuster1997Bidirectional} to generate a set of image features $V = \{v_1,v_2,…,v_k\}, v_i \in \mathbb{R}^D $and a set of word features $ E = \{e_1,…,e_n\} $, $ e_i \in \mathbb{R}^D $ respectively, where $ e_t $ represents t-th word feature. Then we apply attention mechanism to predict the attended image vector $ \mathrm{a}_t^v $  with respect to each word, and calculate the local similarity $ R(e_t, \mathrm{a}_t^v) $ between word feature $ e_t $ and its corresponding attended image vector $ \mathrm{a}_t^v $. Finally, we summarize the local similarity to get the global similarity $ S(I,T) $. The detailed formulas are explained as following:

\begin{equation}
	S_{it} = \dfrac{v_i^T e_t}{\left \| v_i \right \| \left \| e_t \right \|},
\end{equation}

\begin{equation}
	\alpha_{it} = \dfrac{\exp(\tau \overline{s}_{it})}{\sum_{i=1}^{K} \exp(\tau \overline{s}_{it})},
\end{equation}

\begin{equation}
	\mathrm{a}_t^v = \sum_{i=1}^{K} \alpha_{it} v_i,
\end{equation}

\begin{equation}
	R(e_t, \mathrm{a}_t^v) = \dfrac{e_t^T \mathrm{a}_t^v}{\left \| e_t \right \| \left \| \mathrm{a}_t^v \right \|},
\end{equation}

\begin{equation}
	S(I, T) = \dfrac{\sum_{t=1}^{n} R(e_t, \mathrm{a}_t^v)}{n},
\end{equation}

\noindent where $ s_{it} $ represents the similarity between the i-th region and the t-th word. It is normalized as $ \overline{s}_{it} = [s_{it}]_+ / \sqrt{\sum_{t=1}^{n} [s_{it}]_+^2} $ where $ [x]_+ = max(x, 0) $.

\subsection{Training and Objectives}

In order to get different descriptions of the same image, we first train two image caption models separately: Forward LSTM $ L_f $ and Backward LSTM $ L_b $, which generate sentences of positive order and reverse order respectively. After they have been well trained with cross-entropy loss and self-critical optimization on CIDEr \cite{Vedantam2015CIDEr}, we then train an image-text matching model SCAN by employing a hinge-based triplet loss. SCAN model is used to choose the better one which is more alignment to the image between sentences generated by $ L_f $ and $ L_b $.

\noindent\textbf{Training with Cross Entropy Loss.} Given a target ground truth sequence $ y_{1:T} $, the control signal $ \beta $ and the captioning model $ L_f $ with parameters $ \theta $, we minimize the following cross entropy loss:

\begin{equation}
	L_f(\theta) = - \sum_{t=1}^{T} \log(p_{\theta_f}(y_t^* | y_{1:t-1}^*, \beta)),
\end{equation}

\noindent where the value of $ \beta $ depends on the attribute of each ground truth sequence. We will explain the specific setting of $ \beta $ in detail in section 4.2. Similarly, for captioning model $ L_b $, the cross entropy loss is calculated as:

\begin{equation}
	L_b(\theta) = - \sum_{t=1}^{T} \log(p_{\theta_b}(y_t^* | y_{1:t-1}^*, \beta)).
\end{equation}

\noindent\textbf{CIDEr Score Optimization.} After pretrained with cross-entropy loss, captioning model $ L_f $ and $ L_b $ are further trained by reinforce algorithm \cite{Rennie2016Self}. In this stage, parameter $ \beta $ is fixed to an appropriate value according to the needs. The training process is to minimize the negative expected reward:

\begin{equation}
	L_r(\theta) = - \mathbb{E}_{y_{1:T} \sim p_\theta} [r(y_{1:T})],
\end{equation}

\noindent where the reward $ r(\bullet) $ is calculated according to the score of some metric(e.g.CIDEr \cite{Vedantam2015CIDEr}). The gradient can be approximated as:

\begin{equation}
	\nabla_\theta L_r(\theta) \approx - (r(y^s_{1:T}) - r(\hat{y}_{1:T})) \nabla_\theta \log p_\theta (y^s_{1:T}),
\end{equation}

\noindent where $ y^s_{1:T} $ represents a sampled caption and $ \hat{y}_{1:T} $ indicates a result of greedy method.

\noindent\textbf{SCAN training.} Once $ L_f $ and $ L_b $ have been well trained, SCAN is optimized by a triplet loss with margin $ \delta $:

\begin{equation}
	l(I, T) = [\delta - S(I, T) + S(I, \hat{T})]_+.
\end{equation}

Given an image $\mathit{I}$, we first compute the cider score of two sentences generated from $ L_f $ and $ L_b $. One getting higher scores is set as $ T $ and the other is set as $ \hat{T} $.

\begin{table}[h]
	\begin{center}
		\begin{tabular}{|c|c|}
			\hline
			$ \beta $ & tense \\
			\hline
			1 & no v	\\ \hline
            2 & be + v	\\ \hline
            3 & v-ing \\	 \hline
            4 & v \\ \hline
            5 & v-ed	\\ 
			\hline
		\end{tabular}
	\end{center}
	\caption{Specific settings of $ \beta $ for sentence tense.}
\label{Table2}\end{table}

\begin{table}[h]
	\begin{center}
		\begin{tabular}{|c|c|c|c|c|c|c|}
			\hline
			$ \beta $ & B-1 & B-4 & M & R & C & S \\ \hline
			baseline & 77.2 & 36.2 & 27.0 & 56.4 & 113.5 & 20.3	\\ \hline
			1 & 57.6 & 20.1 & 18.7 & 41.7 & 54.1 & 12.6	\\ \hline
            2 & 72.0 & 31.9 & 25.7 & 53.3 & 95.6 & 18.8	\\ \hline
            3 & 77.3 & 37.3 & 28.1 & 57.5 & 117.6 & 21.0	\\ \hline
            4 & 78.3 & 38.0 & 28.1 & 57.8 & \textbf{119.5} & 21.3	\\ \hline
            5 & 78.6 & 38.1 & 28.0 & 57.9 & 119.2 & 21.3	\\ \hline
            6 & 78.6 & 38.0 & 27.8 & 57.8 & 118.3 & 21.1	\\ \hline
		\end{tabular}
	\end{center}
	\caption{Different settings of $ \beta $ for sentence quality. Baseline represents the results of Up-Down model without our control signal. B-1, B-4, M, R, C and S mean BLEU1, BLEU4, METEOR, ROUGE-L, CIDEr and SPICE scores respectively.}
\label{Tablex}\end{table}

\begin{table*}[htbp]
	\begin{center}
		\begin{tabular}{|c|c|c|c|c|c|c|c|c|c|c|c|c|}
			\hline
			 & \multicolumn{6}{c|}{Cross-Entropy Loss} & \multicolumn{6}{c|}{CIDEr Score Optimization} \\
			\hline
			
			Models & B-1 & B-4 & M & R & C & S & B-1 & B-4 & M & R & C & S \\ \hline
			SCST\cite{Rennie2016Self} & - & 30 & 25.9 & 53.4 & 99.4 & - & - & 34.2 & 26.7 & 55.7 & 114 & - \\ \hline
			LSTM-A\cite{8237786} & 75.4 & 35.2 & 26.9 & 55.8 & 108.8 & 20.0 & 78.6 & 35.5 & 27.3 & 56.8 & 118.3 & 20.8 \\ \hline
			RFNet\cite{jiang2018recurrent} & 76.4 & 35.8 & 27.4 & 56.8 & 112.5 & 20.5 & 79.1 & 36.5 & 27.7 & 57.3 & 121.9 & 21.2 \\ \hline
			LBPF\cite{Qin2020Look} & 77.8 & 37.4 & 28.1 & 57.5 & 116.4 & 21.2 & \textbf{80.5} & 38.3 & 28.5 & 58.4 & 127.6 & 22.0 \\ \hline
			AoANet\cite{Huang2019Attention} & 77.4 & 37.2 & \textbf{28.4} & 57.5 & 119.8 & 21.3 & 80.2 & \textbf{38.9} & \textbf{29.2} & \textbf{58.8} & \textbf{129.8} & \textbf{22.4} \\ 
			\hline
			\hline
			Up-Down\cite{Anderson2018Bottom} & 77.2 & 36.2 & 27.0 & 56.4 & 113.5 & 20.3 & 79.8 & 36.3 & 27.7 & 56.9 & 120.1 & 21.4 \\ \hline
			Up-Down+SCAN & 77.1 & 36.4 & 27.6 & 56.6 & 115.5 & 21.0 & 79.2 & 36.9 & 27.8 & 57.1 & 122.6 & 21.3 \\ \hline
			Up-Down+control & 78.2 & \textbf{38.0} & 28.1 & \textbf{57.8} & 119.4 & 21.3 & 80.2 & 37.7 & 28.0 & 57.8 & 125.9 & 21.6 \\ \hline
			Up-Down+control+SCAN & \textbf{78.3} & 37.5 & 28.1 & 57.4 & \textbf{120.3} & \textbf{21.5} & 78.7 & 37.1 & 27.7 & 56.1 & 126.4 & 21.5 \\ \hline
		\end{tabular}
	\end{center}
	\caption{Performance comparisons on MSCOCO Karpathy test split, where B-1, B-4, M, R, C and S represent BLEU1, BLEU4, METEOR, ROUGE-L, CIDEr and SPICE scores respectively. The baseline is \cite{Anderson2018Bottom}. The best results for each metric are marked in boldface.}
\label{Table3}\end{table*}

\section{Experiments}
\subsection{Datasets and Evaluation Metrics}

We use the MSCOCO 2014 captions dataset \cite{Lin2014Microsoft} to evaluate our proposed methods. MSCOCO dataset includes 164,062 images labeled with 5 captions each. Following the Karpathy data split \cite{Karpathy2016Deep} which has been widely used in prior work, we choose 113,287 images for training, 5000 images for validation and 5000 images for test.
We measure the caption quality by using five evaluation metrics, including BLEU-n\cite{papineni-etal-2002-bleu}, ROUGR-L\cite{lin-2004-rouge}, METEOR\cite{Denkowski2014Meteor}, CIDEr\cite{Vedantam2015CIDEr} and SPICE\cite{2016SPICE}.

\subsection{Implementation Details}

Similar to the setting in \cite{Anderson2018Bottom}, we extract image features by employing a Faster-RCNN\cite{Ren2015Faster} model pretrained on Visual Genomes\cite{Krishna2017Visual}. According to the complexity of each image, we extract 10 to 100 ROI(region of interest) pooling vectors whose size is set as 2048. The features extracted from image serve as the input of both captioning model and image-text matching model.

For the captioning model, we set the same hyper-parameters as that in \cite{Anderson2018Bottom}. Specifically, the input word embedding size and both LSTMs’ hidden state size are all set to 1000. The hidden units of the attention module are set to 512. As for the cross-entropy training, we adopt Adam optimizer with the learning rate set as 5e-4, which decays by a factor 0.8 every 3 epochs. The epoch of this training method is set to 35. For self-critical learning, we start from the model which is first trained under cross-entropy training for 30 epochs, and then use SCST training for another 40 epochs with a fixed learning rate of 5e-5.

Here we focus on the setting of the control signal $ \beta $. Whether the $ \beta $ value is set to a continuous value or a discrete value depends on the needs. In order to control the quality of generated sentence, we set $ \beta $ as a continuous value such as cider score between the label sentence and five ground truth sentences. We can also set $ \beta $ to a discrete value when controlling other attributes such as sentence length, sentence tense or number of nouns in the sentence. Specifically, $ \beta $ is set to x when sentence length of x or noun number of x is needed. Setting of $ \beta $ for sentence tense is shown in Table~\ref{Table2}. In the cross-entropy training phase, the value of $ \beta $ is set according to the attributes of each label sentence. In the SCST training and inference phase, $ \beta $ is fixed to an appropriate value according to the demand(4 for sentence quality).

For the image-text matching model, we set the word embedding size to 300 and GRU hidden state size to 1024. The margin $ \delta $ and temperature $ \tau $ are set to 0.2 and 9 respectively. We train SCAN model for 10 epochs with the learning rate set to 5e-4, which decays by a factor 0.8 every 3 epochs.

\begin{table}
	\begin{center}
		\begin{tabular}{|c|c|}
			\hline
			Models & proportion \\
			\hline
			Up-Down & 48.8 \% \\
			Up-Down+SCAN & \textbf{43.1\%} \\				
			\hline
		\end{tabular}
	\end{center}
	\caption{Proportion of poor-quality sentences on MSCOCO Karpathy test split. In this experiment, we define sentences whose cider score is less than 90.0 as poor-quality sentences.}
\label{Table4}\end{table}

\subsection{Quantitive Analysis}

\noindent\textbf{Model Selection with $ \beta $.} To prove that our control signal can determine the sentence quality, we show the performance of baseline Up-Down model with the control signal of different values. With a control signal of sentence quality, results of models by cross-entropy training are reported in Table~\ref{Tablex}. Focusing on the CIDEr score which has become the most important evaluation metric in captioning task, we can observe that higher quality sentences are generated with the increase of $ \beta $. However, when $ \beta $ is set to more than 5, the metric score drops a bit. The phenomenon is probably due to too few sentences scoring more than 5 in the training set. As a result, we select the best result and fix $ \beta $ to 4 in the CIDEr optimization phase and inference phase. The experiments above prove that our control signal can effectively control the quality of captions.

\noindent\textbf{Evaluation of proposed methods.} We report the performance of our proposed methods on MSCOCO Karpathy test split of our proposed method as well as the compared models in Table~\ref{Table3}. Our baseline is Up-Down model proposed in \cite{Anderson2018Bottom}. Up-Down models with SCAN structure or control signal of sentence quality are represented as Up-Down+SCAN and Up-Down+control respectively. Focusing on the CIDEr score in Table~\ref{Table3}, Up-Down+control has a boost improvement compared with baseline, up to 119.4 on CIDEr score with cross-entropy loss. It also improves 4.8 percent on CIDEr optimization. Up-Down+SCAN outperforms the baseline 6.0/5.2 percent with two training methods respectively. Up-Down+control+SCAN combines two proposed methods together, achieving BLEU-4/CIDEr scores of 37.5/120.3 with cross-entropy loss, and 37.1/126.4 with CIDEr optimization. The results with cross-entropy loss surpass the results of other state-of-the-art methods, which demonstrate the effectiveness of our methods.

\noindent\textbf{Reduce the proportion of bad sentences.} We use SCAN model to judge two different sentences generated by captioning model Lf and Lb. It is impossible for a captioning model to perform well on all the images in the dataset. When it generates poor-quality sentences, we need an extra option which could be much better. So the introduction of the SCAN module will reduce the proportion of bad sentences generation. To prove this, we present the proportion of poor-quality sentences generated by Up-Down and Up-Down+SCAN according to the CIDEr score in Table~\ref{Table4}. As the table shows, given a threshold of 90, our proposed method can decrease the proportion of bad sentences effectively, from 48.8$\%$ to 43.1$\%$.

\noindent\textbf{Generality.} To validate the generality of two proposed methods, we further test them in a different situation. For the first method, we take SGAE model \cite{Yang2018Auto} as our baseline. Table~\ref{Table5} shows the performance of SGAE model with the control signal of sentence quality. It can be seen that SGAE+control improves the baseline from BLEU-4: 36.7, CIDEr: 115.7, SPICE 21.0 to 38.2, 120.8, 21.6, which outperforms other state-of-the-art methods in terms of cross-entropy training.

For the second method, we further conduct experiments based on beam search. Beam search is a classic method which can sample multiple captions. It keeps top-k captions ordered by likelihood at each time step and finally chooses one with maximum probability from k captions. However, sentence with maximum probability is not necessarily the best sentence. To solve this problem, we first use beam search method to generate top-k captions, then take SCAN model as an evaluator to judge the quality of top k captions and finally choose one which gets highest similarity score from SCAN model. Table 5 shows the improvements of evaluation metrics when our method is applied to beam search. It is observed that Up-Down+beam+SCAN improves the baseline from BLEU-4: 36.2,CIDEr: 113.5, SPICE 20.3 to 36.5, 116.4, 21.1.

\begin{table}
	\begin{center}
		\begin{tabular}{|m{52pt} < {\centering}|m{16pt} < {\centering}|m{16pt} < {\centering}|m{16pt} < {\centering}|m{16pt} < {\centering}|m{22pt} < {\centering}|m{16pt} < {\centering}|}
			\hline
			Models & B-1 & B-4 & M & R & C & S \\
			\hline
			Up-Down\cite{Anderson2018Bottom} & 77.2 & 36.2 & 27.0 & 56.4 & 113.5 & 20.3 \\
			Up-Down+c & \textbf{78.2} & \textbf{38.0} & \textbf{28.1} & \textbf{57.8} & \textbf{119.4} & \textbf{21.3} \\
			\hline
			SGAE\cite{Yang2018Auto} & 76.8 & 36.7 & 28.1 & 56.9 & 115.7 & 21.0 \\
			SGAE+c & \textbf{78.3} & \textbf{38.2} & \textbf{28.4} & \textbf{57.9} & \textbf{120.8} & \textbf{21.6} \\
			\hline
			\hline
			Up-Down +beam & 77.2 & 36.2 & 27.0 & 56.4 & 113.5 & 20.3 \\
			Up-Down +beam+S & 77.2 & \textbf{36.5} & \textbf{27.9} & \textbf{56.9} & \textbf{116.4} & \textbf{21.1} \\
			\hline
		\end{tabular}
	\end{center}
	\caption{Performance comparisons on MSCOCO Karpathy test split using cross-entropy loss and beam search method. +c and +S represent models with control label and SCAN structure respectively.}
\label{Table5}\end{table}

\begin{table*}[htbp]
	\begin{center}
		\begin{tabular}{|m{22pt}<{\centering}|c|m{130pt}|m{130pt}|m{130pt}|}
			\hline
			\multicolumn{2}{|c|}{image} &
			\vspace{1mm}
			\includegraphics[width=129pt, height=90pt]{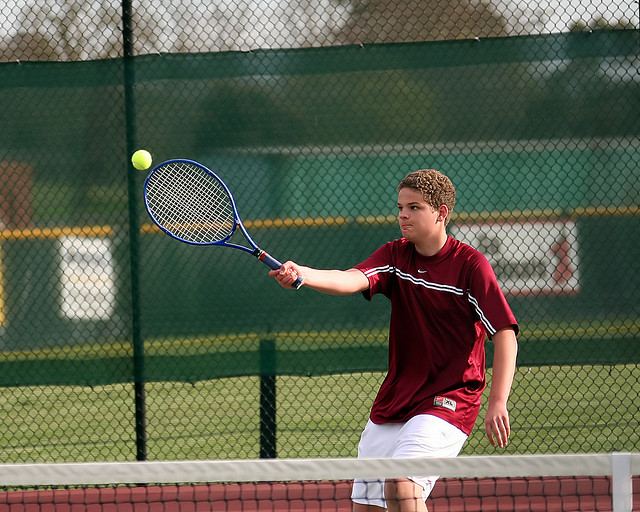} & 
			\vspace{1mm}
			\includegraphics[width=129pt, height=90pt]{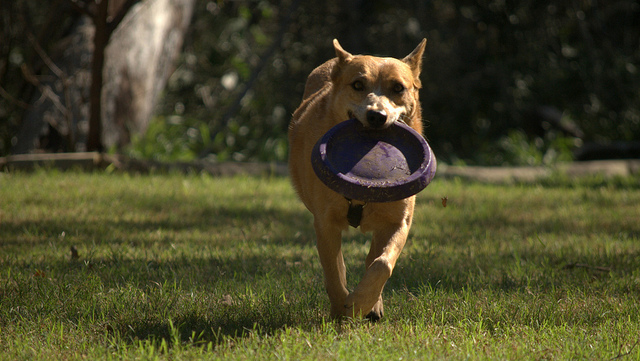} & 
			\vspace{1mm}
			\includegraphics[width=129pt, height=90pt]{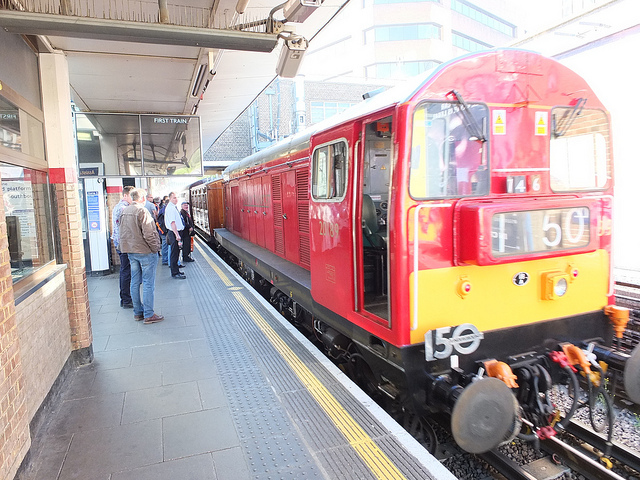} \\
			\hline
		
			& 7 & a tennis player swinging at a ball & a dog holding a frisbee in mouth & a red train pulling into a station \\
			\cdashline{2-5}[2pt/2pt]
			& 8 & a man swinging a racket at a ball & a dog holding a frisbee in its mouth & a red train pulling into a train station \\
			\cdashline{2-5}[2pt/2pt]
			& 9 & a man swinging a tennis racket at a ball & a dog holding a frisbee in its mouth open & a red and yellow train pulling into a station \\
			\cdashline{2-5}[2pt/2pt]
			length & 10 & a man swinging a tennis racquet at a tennis ball & a dog is holding a frisbee in its mouth open & a red and yellow train pulling into a train station \\
			\cdashline{2-5}[2pt/2pt]
			& 11 & a young man swinging a tennis racquet at a tennis ball & a dog holding a frisbee in its mouth in a field & a red and yellow train pulling into a train station platform \\
			\cdashline{2-5}[2pt/2pt]
			& 12 & a young man swinging a tennis racquet on top of a court & a dog holding a frisbee in its mouth in a grassy field & a red and yellow train pulling into a train station with people \\
			\hline
			
			& no v & a person on a court with a tennis racket & a dog with a frisbee in its mouth & a red and yellow train at a train station \\
			\cdashline{2-5}[2pt/2pt]
			& be + v & a boy \underline{\textit{is swinging}} a tennis racket at a ball & a dog \underline{\textit{is holding}} a frisbee in its mouth & a train \underline{\textit{is stopped}} at a train station \\
			\cdashline{2-5}[2pt/2pt]
			tense & v-ing & a young boy \underline{\textit{hitting}} a tennis ball with a racquet & a dog \underline{\textit{holding}} a frisbee in its mouth & a red train \textit{\underline{pulling}} into a train station \\
			\cdashline{2-5}[2pt/2pt]
			& v & a young boy about to \underline{\textit{hit}} a tennis ball & a dog \underline{\textit{carries}} a frisbee in its mouth & a train \underline{\textit{pulls}} up to a train station \\
			\cdashline{2-5}[2pt/2pt]
			& v-ed & a young boy \underline{\textit{dressed}} in red and white playing tennis & a dog \underline{\textit{carried}} a frisbee in its mouth & a red train \underline{\textit{stopped}} at a train station \\
						
			\hline	
			& 2 & a young \textcolor{cyan}{boy} is playing \textcolor{magenta}{tennis} & a \textcolor{cyan}{dog} that is standing in the \textcolor{magenta}{grass} & a red and yellow \textcolor{cyan}{train} pulling into a \textcolor{magenta}{station} \\
			\cdashline{2-5}[2pt/2pt]
			noun's num & 3 & a young \textcolor{cyan}{boy} is playing \textcolor{magenta}{tennis} on a \textcolor{green}{court} & a \textcolor{cyan}{dog} with a \textcolor{magenta}{frisbee} in its \textcolor{green}{mouth} & a red \textcolor{cyan}{train} pulling into a \textcolor{magenta}{train} \textcolor{green}{station} \\
			\cdashline{2-5}[2pt/2pt]
			& 4 & a young \textcolor{cyan}{boy} hitting a \textcolor{magenta}{tennis} \textcolor{green}{ball} with a \textcolor{blue}{racquet} & a \textcolor{cyan}{dog} in a \textcolor{magenta}{field} with a \textcolor{green}{frisbee} in its \textcolor{blue}{mouth} & a red \textcolor{cyan}{train} pulling into a \textcolor{magenta}{train} \textcolor{green}{station} next to a \textcolor{blue}{platform} \\
			\hline
		\end{tabular}
	\end{center}
	\caption{Sample results of controllability with different control signals. In order to fully present the role of each control signal, we separately train three models with different control signal to control the sentence length, sentence tense and number of nouns in the sentence respectively. The results show that for a given image, our method can generate various descriptions according to the demand.}
\label{Table6}\end{table*}

\subsection{Qualitative Result}

\noindent\textbf{Control Sentence Length.} Table~\ref{Table6} shows some examples of images and captions generated by Up-Down\cite{Anderson2018Bottom} model with the control signal. For each image, we set the value of control signal from 7 to 12. As illustrated in Table~\ref{Table6}, we can observe that the generated sentence, which still maintains high-quality, can be accurately fixed to a certain length.

\noindent\textbf{Control Sentence Tense.} A few examples about sentence tense are shown in Table~\ref{Table6}. It is obvious that by controlling the value of control signal, we can generate diverse sentences in any tense we want, which further validates the effectiveness of the control signal.

\noindent\textbf{Control number of nouns.} We employ NLTK tools to mark the part of speech of the sentence. Table~\ref{Table6} shows several examples with different noun quantity requirements on MSCOCO Karpathy test split and verifies the accuracy of our method. It can be seen that with the increase of noun number, objects contained in the description become more comprehensive. For example, for the second image, when we set the value to 2 or 3, the corresponding descriptions lose some details like "frisbee" or "field". By contrast, the caption with 4 nouns is more comprehensive. However, the number of nouns should not be set too much to avoid many useless phrases like "in front of" and "on top of" in the sentences.

\section{Conclusion}

In this paper, two novel methods are introduced to the captioning model. We propose a control signal to enhance the controllability and diversity of existing captioning models. Through control signal, we can control the macroscopic sentence attributes such as sentence quality, sentence length, sentence tense, and number of nouns in the sentence, which enables us to generate diverse sentences according to our needs. We also innovatively train an image-text matching model to judge the quality of sentences generated in forward and backward directions. This method reduces the proportion of poor-quality sentences while improves overall captioning performance.

In terms of cross-entropy training, we achieve a new state-of-the-art performance with Up-Down+control +SCAN. Experiments conducted on the MSCOCO dataset above has demonstrated the generality of our methods. They can be easily applied to most captioning models and other sequence generation task like machine translation.

{\small
\bibliographystyle{ieee_fullname}
\bibliography{egbib}
}

\end{document}